\documentclass[10pt,twocolumn,letterpaper]{article}

\usepackage{cvpr}
\usepackage{times}
\usepackage{epsfig}
\usepackage{graphicx}
\usepackage{amsmath}
\usepackage{amssymb}
\usepackage{algorithm}
\usepackage[noend]{algpseudocode}
\usepackage{algorithmicx}
\usepackage{multirow}
\usepackage[breaklinks=true,bookmarks=false]{hyperref}

\cvprfinalcopy % *** Uncomment this line for the final submission

\begin{document}

%%%%%%%%% TITLE
\title{CoMIC: Good features for detection and matching at object boundaries}

\author{Swarna Kamlam Ravindran\\
Indian Institute of Technology, Madras\\
Chennai, India- 600036\\
{\tt\small ee11s019@ee.iitm.ac.in}
\and
Anurag Mittal\\
Indian Institute of Technology, Madras\\
Chennai India- 600036\\
{\tt\small amittal@iitm.ac.in}
}

\maketitle
%\thispagestyle{empty}

%%%%%%%%% ABSTRACT
\begin{abstract}
Feature or interest points typically use information aggregation in 2D patches which does not remain stable at object boundaries 
when there is object motion against a significantly varying background.
Level or iso-intensity curves are much more stable under such conditions, especially the longer ones.  
In this paper, we identify stable portions on long iso-curves and detect corners on 
them. Further, the iso-curve associated with a corner 
is used to discard portions from the background and improve matching.
Such CoMIC (Corners on 
Maximally-stable Iso-intensity Curves) points yield
superior results at the object boundary regions compared to 
state-of-the-art detectors while performing
comparably at the interior regions as well. This is illustrated in exhaustive matching experiments 
% for point tracking applications on benchmark tracking datasets.
for both boundary and non-boundary regions in applications such as stereo and point tracking for structure from motion
in video sequences.
\end{abstract}

%%%%%%%%% BODY TEXT
\section{Introduction}
Features are the basic building blocks in several tasks in Computer Vision such as
Visual Odometry etc\cite{nister:2004}, 
Structure from Motion (SfM)\cite{sakurada:2013} and Simulataneous Localisation and Mapping (SLAM)\cite{klein:2008}.
% where the latter provides a bound for the minimum eigenvalue in order to ensure that the is well conditioned and above the noise level.
Basic corner detectors or point features popularly used in these applications include
Harris\cite{harris:1988}, Shi and Tomasi \cite{shi:1994} and Hessian\cite{beaudet:1978}
which aggregate image gradients in a patch to find corners.
Many fast variations of point detectors such as 
SUSAN\cite{smith:1997}, AGAST\cite{mair:2010}, FAST\cite{rosten:2006} and FAST-ER\cite{rosten:2008}
have emerged, which perform fast approximations of the gradient computation,
where the latter three use machine learning to train a classifier on a corner model.
The performance of these fast detectors is quite similar to the best-performing point 
detectors, Harris and Hessian\cite{tuytelaars:2004,mair:2010}.
Scale and affine invariant extenstions of these detectors 
\cite{lowe:2004,mikolajczyk:2001,mikolajczyk:2004,agrawal:2008,alcantarilla:2012}
also find use in applications such as Object Recognition and Mosaicing.

\begin{figure}[t]
\begin{centering}
\includegraphics[width=0.236\textwidth]{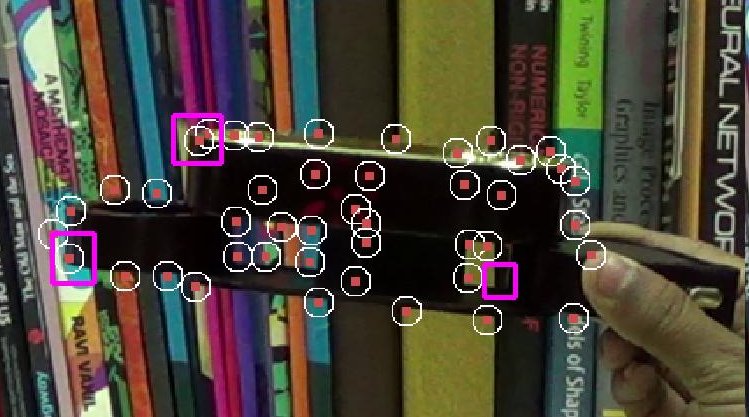} 
\includegraphics[width=0.236\textwidth]{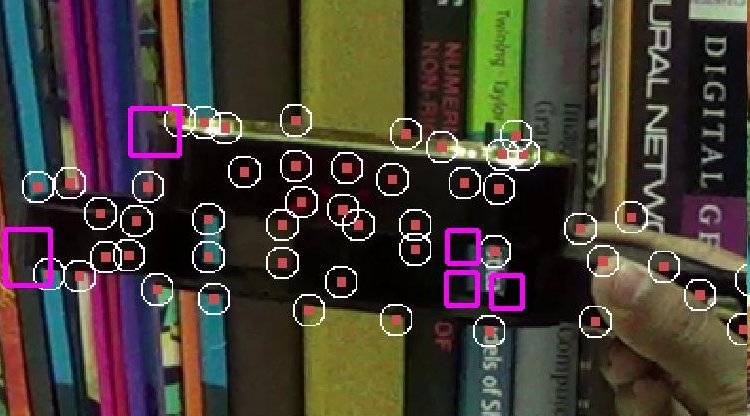} 

\end{centering}
   \caption{Regions shown in boxes correspond to corners in one image that are missed in another or missed in both
   due to the change in gradients associated with a changing background.}
\label{fig:ChangingBound}
\end{figure}

The detection and matching of features across frames of a video is an important first step
in several applications. However, features on object boundaries are typically not utilized in any further
step since the detections and matches are poor in these regions, especially when the object moves against a significantly 
varying background as shown in Fig.\ref{fig:ChangingBound}. % illumination .
This can be attributed to two reasons.
First, the point detectors rely on gradient aggregation in image patches which 
 may span across multiple objects in the scene, leading to errors 
in the boundary region when the object or the camera moves. 
This effect is more pronounced when the background changes and is compounded across frames,
causing a significant drift in the tracks after a number of frames. 
Second, there is a further error introduced in the local template matching stage, where
the correlation values due to varying non-object portions in the patch introduce errors in the 
matching.

\begin{figure}[t]
\begin{center}% \includegraphics[width=0.1\linewidth]{./images/staircase_grad.png} 
\includegraphics[width=0.52\linewidth]{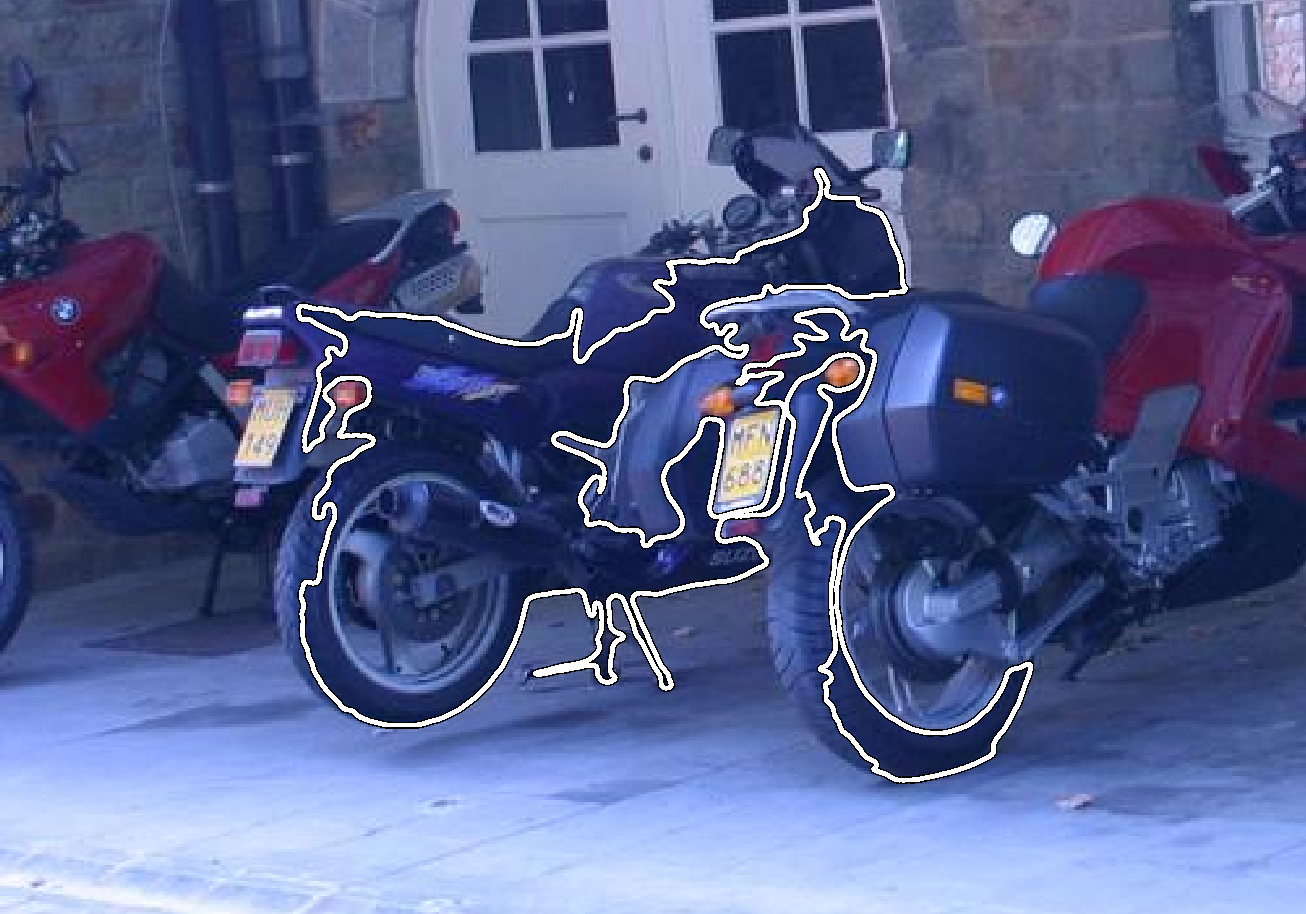}
\end{center}
   \caption{%(a) A typical example of gradient cancellation that occurs due to blur or scale change.
A long iso-curve that forms the boundary of an object.
The information present along such an iso-curve is discarded by MSER.
}
\label{fig:isocurve}
\end{figure}

We address the above problems by proposing a ``corner'' detector on iso-intensity curves. Iso-curves are 
the boundaries of connected components in an image thresholded at a particular
intensity level. 
We note that the boundaries of objects are typically traced by iso-curves which
often move along with the object (Fig.\ref{fig:isocurve}) and can thus be used to detect an object or its parts accurately
even in a changing background.

Our approach to improve the matching accuracy is two-fold.
First, we find points on iso-curves that are more stable and robust to changes in the background,
as compared to points found using patch-aggregation techniques. 
Second, we block out irrelevant background portions of the patch in the template matching stage
using the iso-curve %portion associated with a detected point.
which acts as an effective curve of separation between the object and
the background. 

Features have been detected using iso-curves,
the most popular among them being %iso-curve based detector is 
the Maximally Stable Extremal Regions(MSER) detector\cite{matas:2002}.
MSERs are stable iso-curves that have high Repeatability and Matching scores in image 
matching experiments\cite{mikolajczyk:2005} but return very few detections. 
These may not be sufficient in SfM or point tracking where the overall displacement and 
geometry is drawn from a consensus on corresponding feature points.
The detections are fewer because MSER considers only small, closed iso-curves since features
by definition must be local in order to deal with factors such as occlusion.
This causes MSER to miss information along long iso-curves completely (Fig.~\ref{fig:isocurve}).  
Other approaches have detected corners on iso-curves\cite{cao:2005,perdoch:2007} or edges\cite{tuytelaars:2004}.
However, 
these approaches are again dependent on gradients or use very few points to compute the corner which makes them
quite noisy.  

For matching, the most popular approach is to use gradients distributions\cite{lowe:2004} 
built on the entire patch around the point, which again has problems at the object boundaries
with changing backgrounds.
Shape based descriptions for MSER were proposed by Lowe et al\cite{Forssen:2007}.
While a purely shape based description may be too generic, our
approach effectively combines the use of a curve 
with information from the patch 
as it leads to more distinctiveness of the patch\cite{pinz:2005}.

In this paper, we use the information along long iso-curves and detect corners on portions of them.
We use a measure based on area-change (similar to MSER) for determining 
the local stability of an iso-curve.  Furthermore, we improve the matching using the iso-curve.
% The evaluation of the matching follows that used in VO application.
We demonstrate through extensive visual and quantitative results that such an approach yields corner 
points that perform well on the boundary regions and are therefore useful in 3D Tracking, SfM and 
3D reconstruction applications.

The rest of the paper is organized as follows.  Sec. 2 describes our features.  Sec. 3 contains
the algorithm and some implementation details
in order to efficiently detect such features.  Finally, Sec. 4 presents experimental results
compared to the state-of-the-art detectors on a variety of datasets.

\section{Corner Definition}
We define our feature point such that it satisfies two properties. 
First, it must be found on an iso-curve segment ($ICS$) (a portion of an iso-curve) that remains largely unchanged with respect to intensity perturbations.  Such an $ICS$ is called locally stable in our work.
Second, it must be a corner along the iso-curve according to a measure that evaluates the distribution of the points of the $ICS$ in orthogonal directions.  
We first consider the idea of local stability of an $ICS$.%whose value is higher than a threshold.

\subsection{Local Stability of an Iso-Curve Segment}
We denote an $ICS$ centered at a candidate corner point $p$ on an
iso-curve at intensity $I$, with $k$ points on either side of $p$ as $ICS(I,p,k)$.  
Equivalently, we may denote the $ICS$ as $ICS(I,p,s)$, where $k = c_{scale} \cdot\ s$, $s$ being the scale at 
which the $ICS$ is detected and $c_{scale}$ being a constant. An approximation used in the $ICS$ implementation is described in Sec. 3.

\begin{figure}[t]
\begin{center}
\includegraphics[width=0.77\linewidth]{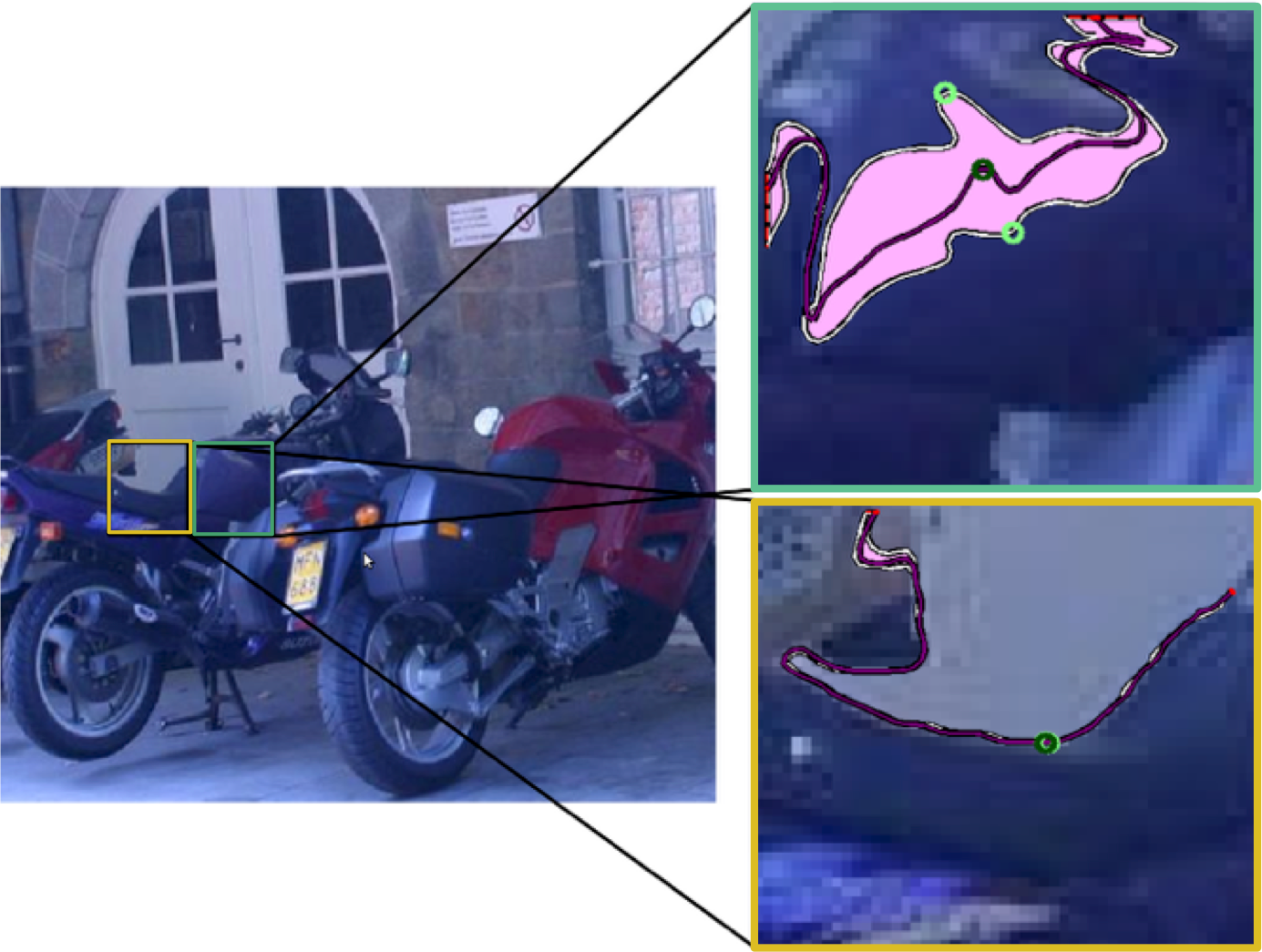} \\%\ \includegraphics[width=0.48\textwidth]{bike1.jpg} \\\\
(a) \hspace{30mm} (b)
\caption{\textrm{Two $ICS$s from image blocks in (a) are shown in red in (b) with the corresponding portions on
their Up and Down $ICS$s at $\delta = 5$ in white 
and their mid-points
% or candidate corner points 
$p$, in green.
The shaded area enclosed between the Up and Down $ICS$s is obtained by connecting their corresponding end-points shown 
in a red dotted line.
While the top row of (b) is an unstable $ICS$, the bottom row shows a stable $ICS$ whose Up and Down $ICS$s are close together.}}
\label{fig:ICS_stability}
\end{center}
\end{figure}

In order to define the stability of an $ICS$, we locate corresponding portions on nearby iso-curves 
at intensities $I+\delta$ and $I-\delta$, which are denoted by $Up(ICS(I,p,s),\delta)$ and $Down(ICS(I,p,s),\delta)$ 
respectively.  
A corresponding portion is identified in an $Up$ or $Down$ $ICS$ by finding points on them that are closest 
to the endpoints of $ICS(I,p,s)$. A few examples are shown diagrammatically in Fig.~\ref{fig:ICS_stability}.
Since iso-­curves do not intersect the $Up$ and $Down$ $ICS$s of an $ICS$ are unique for a particular $\delta$.

The stability
of $ICS(I,p,s)$ can be calculated in terms of a {\em distance measure} between these two open 
curves, $Up(ICS(I,p,s),\delta)$ and $Down(ICS(I,p,s),\delta)$.  
Such a distance measure may be defined by finding corresponding points on the two curves 
and measuring a quantity between some or all of them\cite{perdoch:2007}. These measures can be noisy and
are typically not symmetric.

In this work, we use the area between the two curves
as a measure of stability, as
first proposed by MSER.  While MSER computes this area between two {\em closed curves}, we 
approximate the area $\Delta A$ between two {\em open curve segments}
$Up(ICS(I,p,s),\delta)$ and $Down(ICS(I,p,s),\delta)$
by connecting their end points (Fig.\ref{fig:ICS_stability}(b)).
Such a measure is simple and robust as also seen from the stable performance of MSER.

Given such a variation measure $\Delta A$, we define the stability $\rho$ of $ICS(I,p,s)$ as the inverse of 
its $\Delta A$ divided by its length $len$:
\begin{eqnarray}
    {1 \over \rho(ICS(I,p,s),\delta)} = {\Delta A(ICS(I,p,s)) \over len(ICS(I,p,s))}
    \label{eq:rho_stability}
\end{eqnarray}
%Here $i'$ and $i''$ are the closest points to $i$ on $ICS_{i',k,I+\delta}$ and $ICS_{i'',k,I-\delta}$ respectively.
Essentially, ${1/ \rho}$ measures the average motion of a point on the $ICS$ when the intensity is varied.  
Thus, lower values of ${1/ \rho}$ (or higher values of $\rho$) specify $ICS$s that are relatively stable with respect to intensity variations and can thus be found reliably in 
another image of the same scene. 

We make a further modification to the above measure to make it more robust.  Portions of the curve near the candidate corner $p$ 
are more important than portions away from it.  A large distance between two iso-curve portions near a corner 
must not be averaged to a low distance value due to smaller distances between portions further away from it.
Fig~\ref{fig:ICS_stability}(b) shows such a corner that is unstable, eventhough it has relatively more stable end portions. 
At the same time, a relatively less stable portion far away from the corner should not bring its stability down.

We improve the measure for corner stability by giving a Gaussian-weight to the points in the image 
that are used to compute $\Delta A$ as well 
as the points used to compute $len(ICS(I,p,s))$.  In order to do so in a consistent way, we first assign weights to the points on the iso-curve 
based on their distance along the curve from $p$ using a 1D Gaussian.  While $len$ is calculated from such a weighted curve, the
  $\Delta A$ computation is done by assigning weights to all the points in the 2D image.
Each point in the image is given the weight of the point on the $ICS$ closest to it.  This can be done very fast using the distance transform.  
Such an approach still measures the average motion of a point on the $ICS$ but now does so with a 
Gaussian weight assigned to such points.  

Using this stability measure $\rho$ of $ICS$s,  a non-maximal suppression is done to accurately localize
them. 
% This yields $ICS$s that are locally maximal with respect to $\rho$,
The stability of the $ICS$ is higher than their respective $Up(ICS(I,p,s),1)$ and 
$Down(ICS(I,p,s),1)$, i.e iso-curve segments 
 that are immediately above and below it.  We denote each such \emph{maximally stable 
 iso-curve segment} as $MSICS(I,p,s,\delta)$.

The local maximally stable iso-curve segments thus obtained should be as different from a straight line as possible,
since points on straight lines cannot be localized in another image accurately.
The following approach is used to obtain $MSICS$s of the appropriate shape. 

%%%%%%%%
\subsection{Corners on Iso-Curves}
The second condition we enforce on the feature point is that it must not lie on an $ICS$ that is nearly straight. To detect such distinct and well localized points, we find
corners on the $ICS$. 
A popular concept used to measure the change in the direction at a particular point of the curve is the 
curvature\cite{perdoch:2007,cao:2005}.
However, curvature based methods can be quite sensitive to noise.

\begin{figure}[t]
\begin{center}
\includegraphics[width=0.35\linewidth]{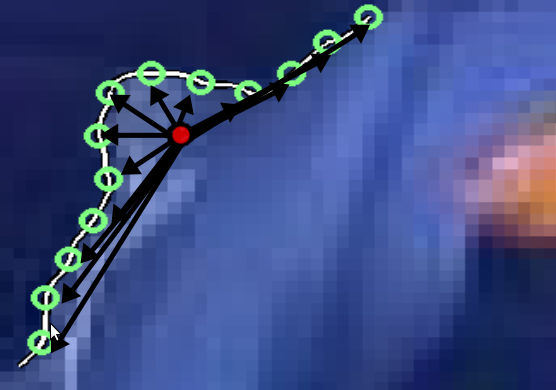}
\end{center}
\caption{
The distribution of points on an $ICS$ w.r.t their mean point (in red) which is used to determine a corner.}
\label{fig:corner}       % Give a unique label
\end{figure}

In this work, we detect a corner by measuring the distribution of points of an $ICS$ centered at a given point (Fig.~\ref{fig:corner}). % using their second moment (covariance) matrix.
A similar technique is used by Tsai et al.\cite{tsai:1999} to find corners on a curve, which was shown to return less spurious corners
compared to the curvature approach.
We compute the covariance matrix $\Sigma$ using:
\begin{equation}
\begin{array}{c}
\begin{split}
&\Sigma_s(t) = G_{0,\sigma} \otimes \\
&\left[ 
\begin{array}{cc}
(x(t)-\bar{x}(t))^2 & (x(t)-\bar{x}(t))(y(t)-\bar{y}(t))  \\
(x(t)-\bar{x}(t))(y(t)-\bar{y}(t)) & (y(t)-\bar{y}(t))^2  
\end{array}  \right]
\end{split}
\end{array}
\end{equation} 

where $t$ is used to index the points on the $ICS$. These points are Gaussian weighted
according to a variance $\sigma$ that is 
proportional to the scale $s$ at which the corner is being detected.

The eigenvalues of $\Sigma$ reflect the distribution of the points of the $ICS$ along two principal orthogonal
directions and high values of both indicate a corner point.  The idea is similar to the Harris Corner 
detector\cite{harris:1988} which works on the second moment matrix of the image gradients.
Several measures have been used in the literature: $det(\Sigma_{s}(t)) - k \cdot trace(\Sigma_{s}(t))^2$\cite{harris:1988},
minimum of the two eigenvalues \cite{shi:1994,tsai:1999},
$det(\Sigma)/trace(\Sigma)$\cite{Brown:2005} and 
$det(\Sigma_{s}(t))/trace(\Sigma_{s}(t))^2$\cite{triggs:2004}.
We use the last measure as it is suitable for point distributions where the number of points on the curve 
is constant. Such a measure is also rotation invariant.
A non-maximal suppression is applied to localize the corner on the iso-curve when 
there are multiple corners in a neighborhood.  

Apart from dealing with the problem of detecting spikes, our approach to find corners on $ICS$s 
has the benefit of not needing exact derivatives. 
This lends the method to fast approximations as explained in the next section.
Compared to traditional 2D corner algorithms, there is a reduction in computation since we work on the 1D curve.
Finally, we define corners as follows:

\paragraph{Definition}
A point  $p$ is said to be a corner at a particular scale $s$ if $ICS(I,p,c_{scale} \cdot\ s)$ is maximally stable
according to the stability measure $\rho$
and is the local maxima of the cornerness measure $\kappa$ along $ICS(I)$ at scale $s$.

An exhaustive search for such maximally stable corner points by investigating
the stability of each segment on each iso-curve present in an image 
would be prohibitively slow.  
We next discuss a method to detect such points efficiently using some approximations.

\section{Algorithm and Implementation Details}
\label{sec:impl}

\begin{figure}[t]
\begin{center}
\includegraphics[width=1\linewidth]{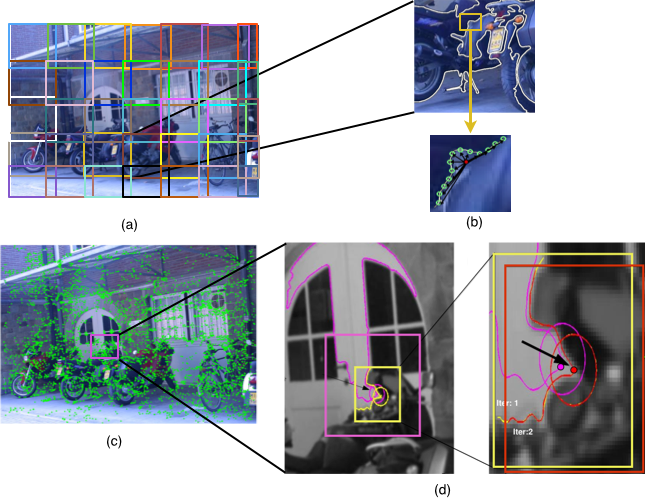} %\ \includegraphics[width=0.48\textwidth]{bike1.jpg} \\
\caption{\textrm{(a) Image divided into blocks of size $f(2s)$. (b) $MSICS$ and Corners found for one sample block. 
For one sample initial corner in (c), (d) shows a block of size $f(s)$ centered on it and the 
redetected $MSICS$ and corner converging
to a more accurate position in two iterations.}}
\label{fig:feature_convergence}
\end{center}
\end{figure}

The first approximation is made in the 
% and the evaluation strategy used to match the detected points consistently.
scale $s$ and stability of an $ICS$ by running an MSER-like algorithm
in an image block as shown in Fig.~\ref{fig:feature_convergence}(d).  The portion of an iso-curve at 
intensity $I$ contained within such a block yields the corresponding 
$ICS(I,p,s)$, where $p$ is the block center and $s$ is related to 
the block size by a constant.

The MSER algorithm that we run in this block is modified in a few ways.
First, we use our stability formulation Eq.\ref{eq:rho_stability} which involves a 
division by the iso-curve length rather than a division by the area of the (closed) iso-curve in 
the standard MSER's stability formulation.
Second, we calculate the areas and the curve lengths using a Gaussian weight on the image points as explained before.  
The Gaussian weighting also ensures that the blocking has negligible effect on the accuracy of the method 
since the points near the block edges will have low weights.  
Furthermore, since we have to compute the measure only in a small neighborhood 
of $ICS(I,p,s)$ (between $Up(ICS(I,p,s),\delta + 1)$ and $Down(ICS(I,p,s),\delta - 1)$), we can 
use a region-growing algorithm for MSER computation\cite{nister:2008}. It is linear in the number
of pixels used in computing the MSER, which is a very small number of pixels in the 
neighborhood of the $ICS$, rather than the whole block.  

A second approximation is used in convolutions/summations for cornerness calculation, where the 1D Gaussian 
function is replaced by an average filter that runs through
multiple iterations to yield approximately the same result due to the Central Limit Theorem. 
The idea is similar in spirit to the approximate 2D Gaussian implemented in SURF\cite{bay-tuytelaars-vangool:06} and is much 
faster due to the use of Dynamic
Programming.  Such an approximation is possible as our cornerness measure is much more robust to weight errors as opposed to other possibilities such as the curvature.

Finally, we describe a two-stage approach that reduces the number of points that have to be analyzed.
We detect initial corners at scale $2\cdot s$ and obtain corners desired at scale $s$ from them through
an iterative procedure.
In the first {\em Initialization} stage, we detect corners $C$ at scale $2\cdot s$ using 
blocks of size BxB (Fig. \ref{fig:feature_convergence}(a)). No weights are used in this stage.
For an image of size MxN and a shift of $B \over 2$ there 
are ${2M \over B}\cdot{2N \over B}$ overlapping blocks. Since the $MSICS$ computation is linear in the
number of pixels in the block, the time for initial $MSICS$ computation $t_{init}$ remains
at $O(M \cdot N)$.

Assuming that such stable iso-curves at a higher scale do not change drastically 
when the scale is reduced by half, the {\em Feature Convergence} procedure proceeds by centering
a weighed block of size $B \over 2$x$B \over 2$ on the corner detected at $2s$, as 
shown in \ref{fig:feature_convergence}(d)).
The modified local MSER and corner detection algorithms are applied on it again.
% local stability more for smaller block
If the redetected corner $c_{new}$ does not change, it is taken to be a maximally stable feature point. 
However, in case the corner shifts to a new point on a nearby
$ICS$, a weighted image block of the same scale $s$ is centered at $c_{new}$ and used to redetermine
the $MSICS$ and the nearest corner point on it, and the process is iterated. 
The fixed point of this iteration, 
if present, yields a point that satisfies both the conditions for our corners.  

The $MSICS$ detection takes $B^2 \over 4$ computations on this smaller block.
For $C$ initial corners and $k$ iterations per corner the convergence 
takes $k \cdot C\cdot B^2 \over 4$ operations.
Typically, $C \ll M\cdot N$ and the average value of $k$ was experimentally
found to be $\leq 3$, therefore the time complexity for 
convergence is $O(B^2)$ and the total time complexity is $O(M \cdot N)$.

The whole iterative 
procedure is illustrated in Fig.~\ref{fig:feature_convergence} while Algorithm~\ref{fig:algorithm1} describes 
the entire algorithm.

It is to be noted that the purpose of using a larger scale during {\em Initialization} is only 
to reduce the computation time. The {\em Feature Convergence} stage ensures that the detected corner is in
the center of the block so that the approximations used in the $ICS$ calculation are consistent.
The desired scale $s$ is taken to be 8.4 and $B$ to be 100, so that each final $ICS$ has about 25 points.

\makeatletter
\def\BState{\State\hskip-\ALG@thistlm}
\makeatother
\algdef{SE}[DOWHILE]{Do}{doWhile}{\algorithmicdo}[1]{\algorithmicwhile\ #1}%

\begin{algorithm}
\caption{Iterating the initialized corners to convergence}\label{fig:algorithm1}
\begin{algorithmic}[1]
\Procedure{Feature Extraction}{}
\State $\textbf{Input} \gets \text{Image $I$, Scale $s$, Delta $\delta$}$
\State $\textbf{Ouput} \gets \text{Corners $C_{new}$}$
\BState \emph{Initialization}:
\State Get Blocks $B \text{ from } I \text{, where block size = $f(2s)$}$ 
\State \begin{tabular}{cl} Compute the set $M$ of $MSICS(I,p,2s,\delta/2)$s, \\
$\forall \  b \in B \text{ using Eq.}$~\ref{eq:rho_stability}
       \end{tabular}
\State $\text{Determine Corners $C$ } \forall m \in M $ %\text{ according to Eq.}$~\ref{eq:kappa_cornerness} 
\For{$c \in C$}
% \BState \emph{Convergence loop}:
\State $m_{old} \gets MSICS(I,p,2s,\delta/2) \text{ and } c_{old} \gets c$
\State Corner movement $d = 100$
\While{$d > 0$}
\State $b_{new} \gets \text{block of size } f(s) \text{ centered at } c_{old}$ 
\State \begin{tabular}{cl} $m_{new} \gets MSICS(I,c_{old},s,\delta)$ redetected\\
in $b_{new}$ s.t $m_{new}$ is similar in shape to $m_{old}$\\
\end{tabular}
% \If{$Shape Similarity(m_{new},m_{old}$) is high}
\State $\text{ Redetect corner $c_{new}$ on $m_{new}$}$
\State $d = dist(c_{old},c_{new})$
\State $m_{old} \gets m_{new} \text{ and } c_{old} \gets c_{new}$
% \Else { Return}
% \EndIf
\EndWhile
\EndFor
\EndProcedure
\end{algorithmic}
\end{algorithm}

\section{Matching Strategy}
% \label{sec:matchStrat}

\begin{figure}[t]
\begin{center}
\includegraphics[width=1\linewidth]{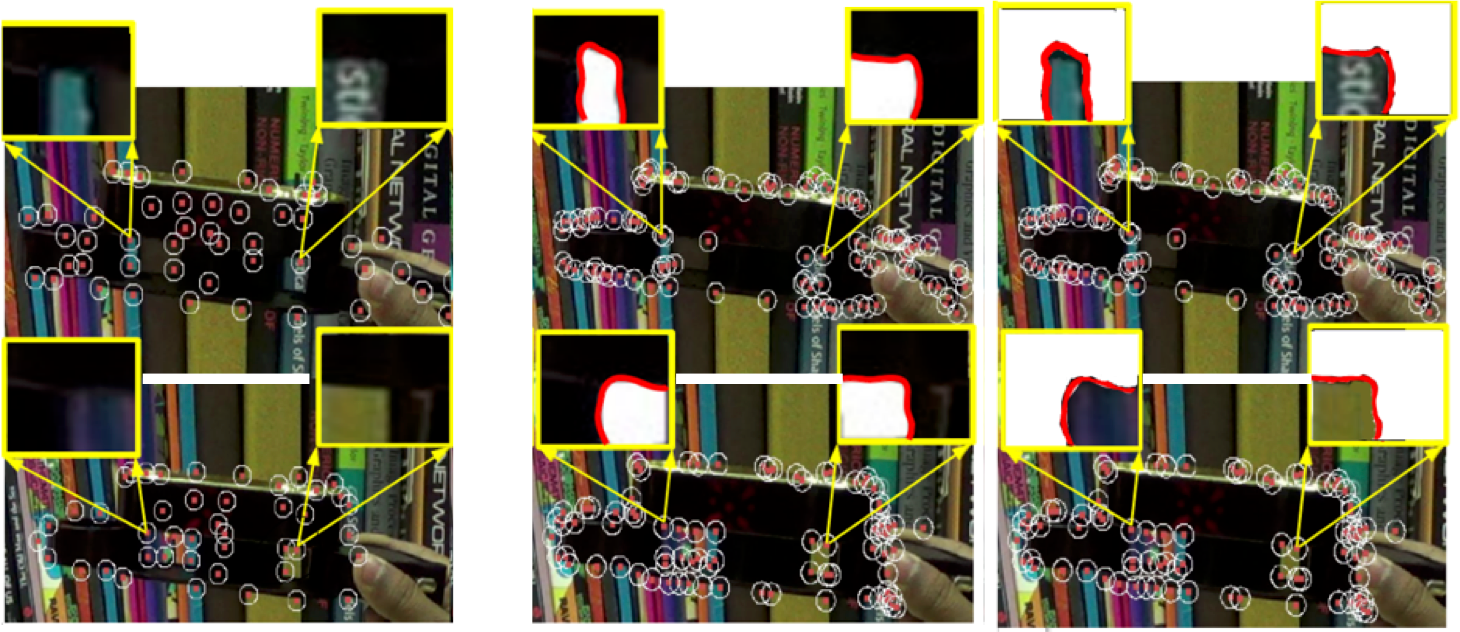} \\
(a) \hspace{20mm} (b) \hspace{30mm} (c)
\caption{\textrm{Regions used for SSD computation (a) Harris (b) CoMIC $ic+$ regions (c) CoMIC $ic-$ regions }}
\label{fig:matching}
\end{center}
\end{figure}
% It is assumed in point tracking that scale and viewpoint changes between consecutive 
% frames is small. This assumption on the camera movement enables us to use a simple 
A simple translational motion model is used to match points in a video sequence, similar to the protocol
followed in Visual Odometry\cite{nister:2004}. This is sufficient due to the low time gap between
consecutive frames. Each point in the current frame is assumed to have moved to a 
location that is within a radius $r$ (dependent on the scale $s$ used to detect the point) 
of that point in the next frame.
The value of $r =20$ pixels was sufficient for our experiments.

We extract a patch $P_{f}$ as a 23x23 window centered on the candidate feature point.
Matching is done using a simple Sum-of-squared-differences (SSD) of the appearance 
of the two patches. 
A simple SSD based matching is sufficient\cite{Baker:2004} in tracking applications although more
complicated invariant descriptors can be used for more complex matching tasks using a similar framework.

% The extremal regions $R$ used in CoMIC features to obtain  $Boundary(R)$ 
$MSICS$s are used as the boundary 
to separate the regions on either side of the $MSICS$, which are denoted
as $ic+$ and $ic-$ respectively.
This is shown pictorially in Fig.\ref{fig:matching}(b), where
they correspond to
Background (BG)/ Foreground (FG) regions for patches on the object boundary.
Since the FG and BG regions are not known in advance, the two sides are matched separately and the 
minimum of their match distances $d_{min}$ is taken to be the match score for that
feature patch. 
% Shape descriptors (mention Shape context?) obtained from the shapes of the $MSICS$s from the two patches are used as an additional check. 
The steps in Matching are listed in Algorithm\ref{fig:algorithm2}, where steps
$8-10$ are unique to our matching approach. The other detectors compute the SSD
between the two full patches as they have no FG/BG separation technique (Fig~\ref{fig:matching}(a)).

\begin{algorithm}
\caption{Matching}\label{fig:algorithm2}
\begin{algorithmic}[1]
% \Procedure{Feature Extraction}
\State \begin{tabular}{cl} $\textbf{Input} \gets \text{Images $I_1, I_2$, Features $F_1, F_2$,scale $s$,}$\\
 \text{radius $r$, $ICS$s $ICS_1$,$ICS_2$, threshold $thresh$}
\end{tabular}
\State $\textbf{Output} \gets \text{Matches} M$
% \State $d_{min} = highVal$
\For{$f1 \in F1$}
\State $P_{f1} \gets $ patch around $f1$
\State $F2_{close} = \{f2 \mid ||f2 - f1|| < r\}$ 
\For{$f2 \in F2_{close}$}
\State $P_{f2} \gets $ patch around $f2$

\State Obtain $P_{f1}^{ic+}, P_{f1}^{ic-}$  from $P_{f1}$ and $ICS1$ 
\State Obtain $P_{f2}^{ic+}, P_{f2}^{ic-}$ from $P_{f2}$ and $ICS2$

\State \begin{tabular}{cl} $d_{min} = $  & $ min(ssd(P_{f1}^{ic+},P_{f2}^{ic+}),ssd(P_{f1}^{ic+},P_{f2}^{ic-}),$  \\
& $ssd(P_{f1}^{ic-},P_{f2}^{ic+}), ssd(P_{f1}^{ic-},P_{f2}^{ic-}))$ 
\end{tabular}
\If {$d_{min} < thresh$}
\State $match(f1) = f2$
\EndIf

\EndFor
\EndFor
% \EndProcedure
\end{algorithmic}
\end{algorithm}

\section{Experimental Results}
We demonstrate the effectiveness of our technique on a variety of videos where the object moves
in a changing background, using the SSD-based template matching technique for point features as described above.

\subsection{Experimental Setup}
\emph{Datasets:}
The performance of some detectors on 3D objects has been evaluated in \cite{Fraundorfer:2004,Moreels:2005}.
% Detectors perform well on these datasets even at the object boundary regions since 
The changes in the background are negligible in these controlled environments, and often the background itself is 
homogeneous.
Furthermore, there is no means to analyse the performance of the feature points lying 
at the object boundary.

In order to evaluate the performance of point detectors at the object boundary regions and under a 
varying background, we have designed the challenging CoMIC dataset
which has objects, homogeneous and textured, moving against a set of differently textured books.
With the knowledge of the static background image, the delineation of the foreground object boundary
is made possible through background subtraction.
This enables an analysis of the performance of the features at the 
boundary and non-boundary regions of the object.

We also show results on sequences from the Middlebury stereo dataset\cite{scharstein:2002}, 
where the object boundary regions are affected by parallax and motion against a textured background.
% that forms the benchmark for testing 3D reconstruction algorithms does not have

We also demonstrate the overall effectiveness of our approach through experiments on 
a subset of sequences from popular tracking datasets such as 
KITTI Vehicle dataset\cite{geiger:2013}, PROST\cite{Santner:2010},
VoT\cite{Kristan:2013} and Cehovin\cite{cehovin:2011}  
which have mostly rigid objects moving against a changing background,  
samples and description of which are given in the supplementary section.
The evaluation is done on the features on the object alone, obtained after the background is
subtracted out in the case of the CoMIC dataset, using the groundtruth depth
discontinuity map in the case of Middlebury sequences and approximated using the groundtruth 
bounding box for the other datasets.
All the experiments were run on images resized to a height of 700 pixels.

\emph{Detectors compared:}
Out of the point detectors in the literature, we compare the performance of our approach with Harris, Hessian 
and FAST-9 (performance is quite similar to FAST-ER\cite{rosten:2008} and AGAST\cite{mair:2010}) detectors that have been found to perform best in 
comparative studies\cite{tuytelaars:2008,rosten:2008}.
Experiments against these detectors were performed 
using codes from \cite{HarrisHessianCode} and \cite{fastercode} respectively.
We do not compare with scale and affine-invariant detectors (Harris-affine and Hessian-affine) since 
these do not perform as well as basic
detectors in these tasks, where there is no significant scale or affine variations between consecutive frames.
However, we compare with MSER 
since their method is closely related to ours.

\emph{Parameters:} 
% Using the default cornerness thresholds for these detectors results in very few detections.
For a fair comparison, we equalize the number of detected features on the object across detectors.
We vary the threshold of each detector to get the same number of points in the first frame of the sequence 
in a manner similar to the 
evaluation in CenSurE\cite{agrawal:2008}.
For this value of the threshold, features are 
 detected on all other images in the sequence.
 In cases where the detector returns very few points in general,
the threshold is fixed to yield the maximum number of points it can return.
For CoMIC, only the stability value $\rho$ is varied to obtain a given number of points.

\emph{Evaluation Criteria:}
The Matching Score $M_{score}(i)$ is obtained as the ratio of the Number of matches $M_{i}$ 
in the $i^{th}$ frame and the Number of detections $N_{i-1}$ in the $(i-1)^{th}$ frame. 
% \begin{eqnarray}
%     M_{Score_i} = {M_{i} \over min(M_{0_i},M_{0_{i+1}})}
% \end{eqnarray}
This measure can be extended to obtain a score for the Number of residual Matches 
$ResM_{i,n}$ in the $i^{th}$ frame that match consistently across $n$ frames,
when the detection is done in the $(i-n)^{th}$ frame.
% compared to the points matched in the $(i-n)^{th}$ frame.

\begin{eqnarray}
    M_{score}(i,n) = {ResM_{i,n} / N_{i-n}} %min(M_{0_i},M_{0_{i-n}})}
    \label{eq:match}
\end{eqnarray}

Thus, out of $N_{i-n}$ points detected in frame $(i-n)$, $ResM_{i,n}$ points have matches
in each of the $n$ frames in between. 
Such a normalization gives a quantitative indication of the 
resilience of the detector, which determines the number of points that can reliably be
tracked over a number of frames. 
This corresponds to resilience to the changes in the background 
for the sequences used in the experiment.
It also gives an 
idea about the interval one needs to choose to redetect points in the frames.

\emph{Groundtruthing:}
Matches that are stable for $n=5$ were taken to be true matches, since it is highly
unlikely that these points match incorrectly in all five frames.
The error due to false matches was found to be about 10\% which becomes negligible
for larger values of $n$.
Duplicate matches are removed by one-to-one matching.

\subsection{Results}
\begin{figure}
\begin{centering}
\includegraphics[width=1\linewidth]{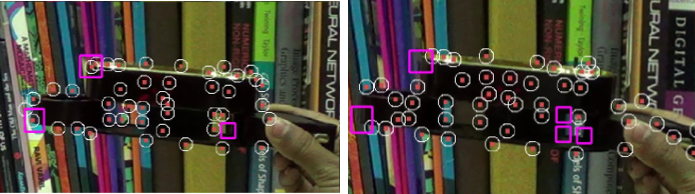} \\
(a) Harris \\
\includegraphics[width=1\linewidth]{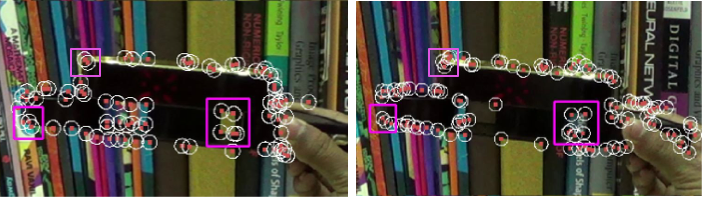} \\
(b) CoMIC \\
\end{centering}
   \caption{CoMIC misses fewer points on the object boundary than Harris as seen from the regions in the colored boxes.
   %    {\bf Show the CoMIC point along with the part on which ICS is detected as it is tracked.}
} 
\label{fig:bound_track}
\end{figure}

\subsubsection{Results on CoMIC dataset}
\begin{table*}[!ht]
\begin{center}
\scriptsize
\begin{tabular}{ |c|c|c|c|c|c|c|c|c|c|c|c|c|c|c|c|c| }
\hline
\multirow{2}{*}{Object}&\multirow{2}{*}{Sequence}&
\multicolumn{5}{|c|}{Boundary} &
\multicolumn{5}{|c|}{{Non-Boundary}} &
\multicolumn{5}{|c|}{{Overall}}\\
 \cline{3-17}
%  &&MS (\%)  & \# &MS (\%)& \#&MS (\%)& \#&MS (\%)& \#&MS (\%)& \# \\
 &&CoMIC& Har &Hes & Mser& Fast &CoMIC&Har &Hes & Mser& Fast &CoMIC& Har &Hes & Mser& Fast \\
\hline \hline
\multirow{4}{*}{\parbox{1.2cm}{Textured}} & Pens & {\bf 40.9} & 24.7 & 23.4 & 9.8 & 13.3 & {\bf 50.8} & 45.8 & 43.5 & 30.9 & 36.1 & {\bf 46.2} & 37.8 & 35.5 & 25.9 & 27.1 \\
\cline{2-17} & Doll &  {\bf 34.6} & 19.5 & 18.2 & 11.3 & 12.0 & 44.9 & {\bf 46.5} & 44.0 & 32.1 & 28.6 & {\bf 40.8} & 36.1 & 34.8 & 27.9 & 20.4  \\
\cline{2-17}& Toy & {\bf 28.4} & 21.6 & 18.5 & 12.6 & 13.0 & {\bf 38.7} & 37.7 & 38.2 & 28.5 & 25.6 & {\bf 34.8} & 31.5 & 31.8 & 25.0 & 21.1\\
\cline{2-17}& Hero & {\bf 39.2} & 27.0 & 24.5 & 12.6 & 16.6 & 45.9 & {\bf 46.8} & 43.6 & 29.2 & 28.4 & {\bf 43.9} & 41.1 & 39.2 & 27.1 & 25.3  \\
\cline{2-17}& Race-car & {\bf 37.0} & 23.2 & 22.7 & 11.9 & 12.8 & 49.9 & {\bf 53.2} & 48.5 & 32.4 & 36.3 & {\bf 44.7} & 42.3 & 40.6 & 29.3 & 29.4 \\
\hline \hline
\multirow{4}{*}{\parbox{1.2cm}{Homogeneous}} & Box & {\bf 36.5} & 33.1 & 28.2 & 10.3 & 20.5 & {\bf 53.3} & 47.0 & 30.2 & 35.5 & 23.7 & {\bf 40.1} & 38.6 & 29.2 & 25.3 & 21.8 \\
\cline{2-17}& Tape-Box & {\bf 47.6} & 31.9 & 26.5 & 10.5 & 16.9 & {\bf 56.9} & 37.7 & 37.3 & 25.5 & 25.8 & {\bf 51.2} & 35.3 & 31.5 & 20.9 & 19.7 \\
\cline{2-17}& House & {\bf 37.2} & 21.5 & 24.4 & 8.7 & 15.8 & 39.9 & {\bf 41.1} & 32.9 & 30.0 & 19.1 & {\bf 38.8} & 35.0 & 29.3 & 26.3 & 17.0  \\
\hline \hline 
\multirow{2}{*}{\parbox{1.1cm}{Stereo}} & Tsukuba & {\bf 32.6} & 20.9 & 29.2 & 11.6 & 30.7 & 48.2 & 47.8 & {\bf 51.1} & 43.3 & 43.9 & {\bf 81.1} &68.4 & 79.7 & 54.9 & 74.2 \\
\cline{2-17}& Cones & {\bf 22.1} & 10.1 & 19.6 & 4.4 & 13.5 & 35.1 & {\bf 44.1} & 37.8 & 43.1 & 32.2 & {\bf 57.4} & 53.3 & 56.1 & 47.5 & 44.9 \\
\hline \hline 

\end{tabular}
\end{center}
\caption{Average Matching Score $M_{score}(5)$ on the CoMIC dataset}
\label{tab:score}
\end{table*}

\begin{table*}[ht]
\begin{center}
\scriptsize
\begin{tabular}{ |c|c|c|c|c|c|c|c|c|c|c|c|c|c|c|c|c| }
\hline
\multirow{2}{*}{Object}&\multirow{2}{*}{Sequence}&
\multicolumn{5}{|c|}{Boundary} &
\multicolumn{5}{|c|}{{Non-Boundary}} &
\multicolumn{5}{|c|}{{Overall}}\\
 \cline{3-17}
%  &&MS (\%)  & \# &MS (\%)& \#&MS (\%)& \#&MS (\%)& \#&MS (\%)& \# \\
  &&CoMIC& Har &Hes & Mser& Fast &CoMIC&Har &Hes & Mser& Fast &CoMIC& Har &Hes & Mser& Fast \\
\hline \hline
\multirow{4}{*}{\parbox{1.2cm}{Textured}}
& Pens & {\bf 27.5} & 5.7 & 15.5 & 2.6 & 6.4 & 45 & 20.1 & {\bf 45.1} & 28.8 & 32.5 & {\bf 72.5} & 25.8 & 60.7 & 31.5 & 38.9 \\
\cline{2-17} & Doll & {\bf 36.4} & 7.0 & 19.7 & 5.5 & 15.2 & 78.2 & 27.7 & {\bf 90.5} & 61.7 & 40.6 & {\bf 114.6} & 34.7 & 110.2 & 67.2 & 55.8  \\
\cline{2-17}& Toy &  {\bf 23.1} & 5.0 & 13.6 & 4.4 & 9.4 & 56.6 & 16.3 & {\bf 63.1} & 38.7 & 39.4 & {\bf 79.7} & 21.4 & 76.7 & 43.1 & 48.8 \\
\cline{2-17}& Hero & {\bf 38.7} & 8.5 & 19.1 & 5.7 & 15.5 & 106.9 & 36.4 & {\bf 110.0} & 76.0 & 79.1 & {\bf 145.6} & 44.9 & 129.1 & 81.7 & 94.7  \\
\cline{2-17}& Race-car & {\bf 40.5} & 7.9 & 20.5 & 4.1 & 11.4 & 87.2 & 32.6 & {\bf 101.7} & 65.1 & 84.3 & {\bf 128.8} & 40.7 & 122.2 & 69.2 & 95.7  \\
\hline \hline
\multirow{4}{*}{\parbox{1.2cm}{Homogeneous}} & Box & {\bf 27.3} & 9.7 & 21.1 & 3.5 & 17.9 & 14.2 & 11.7 & 15.9 & {\bf 16.8} & 10.4 & {\bf 41.5} & 21.4 & 37.0 & 20.3 & 28.2 \\
\cline{2-17}& Tape-Box & {\bf 33.1} & 9.6 & 16.1 & 2.8 & 15.4 & {\bf 21.8} & 14.5 & 15.6 & 14.4 & 9.2 & {\bf 54.9} & 24.0 & 31.7 & 17.2 & 24.6\\
\cline{2-17}& House &{\bf 26.7} & 6.3 & 21.1 & 3.5 & 19.0 & 49.2 & 28.4 & 43.9 & {\bf 63.5} & 21.0 & {\bf 75.9} & 34.7 & 65.0 & 67.0 & 40.0 \\
\hline \hline 
 \multirow{2}{*}{\parbox{1.2cm}{Stereo}} & Tsukuba &{\bf 278.0} & 112.0 & 266.0 & 56.0 & 171.0 &  503.0 & 489.0 & 514.0 & {\bf 544.0} & 409.0 & {\bf 881.0} & 591.0 & 763.0 & 599.0 & 571.0 \\
\cline{2-17}& Cones & {\bf 338.0} & 188.0 & 299.0 & 121.0 & 330.0 & 514.0 & 429.0 & {\bf 524.0} & 450.0 & 471.0 & {\bf 852} & 614.0 & 817.0 & 571.0 & 797.0 \\
\hline \hline 

\end{tabular}
\end{center}
\caption{Average Number of matches $ResM_5$ on CoMIC dataset and Middlebury stereo pairs}
\label{tab:matches}
\end{table*}

Quantitative scores for the resilience of the 
feature are computed using Eq.\ref{eq:match} with $n=5$. The Matching score 
over 5 frames, $M_{score}(5)$ and the Number of matches that survive through 5 frames, $ResM_5$
are computed for every fifth frame and averaged over all the frames in the dataset.
%include lukas kanade in this.
Scores are shown separately for points on the 
object boundary, internal points and all the points for the average $M_{score}(5)$ on the CoMIC dataset 
in Table\ref{tab:score} and $ResM_5$ in Table\ref{tab:matches}.

CoMIC generally outperforms the state-of-the-art detectors, 
in terms of both $M_{score}$ and $ResM$
at the boundary regions of both homogeneous as well as textured objects 
moving against a textured background. 
It performs comparably at non-boundary regions, where it closely
follows Hessian and Harris, to yield the best scores on the full object. 

CoMIC yields substantially more matches with high resilience in the boundary regions and overall, starting 
with approximately the same number of
detections on the object as others in the initial frame. 
This is useful in applications where a high number of matches increases the
consensus on the pose of the object, especially at the boundaries.
This is seen in 
Fig~\ref{fig:bound_track} where several boundary points missed by Harris are 
detected by CoMIC. While these points are also correctly matched in CoMIC, 
very few points from the gradient based detectors match at the boundary regions. These are shown in videos
attached with the supplementary section.

The superior performance at the boundaries can be attributed to 
iso-curves being relatively unaffected by the change in the gradients 
at boundaries when the object moves with respect to its background. 
The feature patch is treated as a whole in gradient based methods, 
causing a matching failure when the background portion in the patch changes.
CoMIC's $MSICS$ acts as a reliable segmentation of the object in the neighborhood of that point.
Examining regions on either side of it separately 
ensures that a matching portion is consistently associated with FG or BG as shown
in Fig.\ref{fig:matching}. This reduces the 
ambiguity in matching and results in higher matching scores.
Such a technique may also be incorporated more generally into a sophisticated descriptor built 
using image intensities on each side of the separation boundary for other applications.
It may also lead to better learning and discrimination of the FG portions in object tracking\cite{grabner:2007}.

Harris and Hessian mostly perform well in the internal object regions and especially when it is textured,
while MSER and CoMIC perform well in homogeneous internal regions with distinctive boundaries.
FAST, while being the fastest detector, yields less than remarkable scores in the comparison.
Internal points are not as affected by changes in the background
and therefore information from the entire patch used in gradient based detectors benefits the matching.
On the other hand, information lost on one side of the curve costs our performance slightly for internal features.
However, apart from being useful in boundary regions, such an approach may even help
non-boundary regions in the case of partial occlusion.

\subsubsection{Stereo Matching}
We observe similar results in the case of stereo matching on
Tsukuba and Cone stereo pairs in the Middlebury dataset.
The features evaluated with groundtruth information in terms of $M_{score}(1)$ and $ResM_1$ 
are shown in Table\ref{tab:score} and Table\ref{tab:matches}.
Again, the spatial windows used for gradient aggregation in Harris etc are affected when the
windows span multiple objects in the scene.

\subsubsection{KITTI and other datasets}
\begin{table*}[ht]
\begin{center}
\scriptsize
\begin{tabular}{ |c|c|c|c|c|c|c|c|c|c|c| }
\hline
\multirow{2}{*}{Sequence}&
\multicolumn{5}{|c|}{Matching Score} &
\multicolumn{5}{|c|}{{Number of Matches}} \\
 \cline{2-11}
%  &&MS (\%)  & \# &MS (\%)& \#&MS (\%)& \#&MS (\%)& \#&MS (\%)& \# \\
&CoMIC& Harris&Hessian& MSER&FAST&CoMIC& Harris& Hessian&MSER & FAST\\
\hline \hline
Board& {\bf 49.6} & 42.7 & 40.1 & 26.4 & 24.5 & {\bf 111.1} & 41.9 & 85.9 & 59.3 & 51.6\\ 
\hline
Lemming&  39.2 & {\bf 43.3} & 35.8 & 27.2 & 14.0 & {\bf 30.6} & 14.6 & 29.3 & 18.0 & 5.4 \\
\hline 
Box& 35.5 & 35.9 & {\bf 36.9} & 31.7 & 19.1 & {\bf 32.5} & 11.8 & 28.7 & 23.7 & 12.0  \\
\hline \hline
Cycle& {\bf 28.9} & 24.1 & 26.3 & 17.0 & 11.6 & {\bf 45.7} & 7.8 & 43.2 & 17.3 & 16.9 \\
\hline
Cup& 44.0 & {\bf 45.1} & 37.9 & 23.9 & 14.2 & {\bf 32.1} & 14.5 & 27.7 & 16.5 & 4.1 \\
\hline \hline
Can& {\bf 27.6} & 18.7 & 18.3 & 15.4 & 9.8 &{\bf 42.2} & 10.9 & 26.9 & 23.6 & 11.9 \\
\hline
Dino& 39.4 & {\bf 39.8} & 38.1 & 28.8 & 17.7 & {\bf 87.6} & 47.9 & 77.1 & 71.9 & 22.7 \\
\hline \hline
Car-A& 18.8 & {\bf 21.5} & 16.3 & 12.1 & 9.1 & {\bf 65.1} & 45.4 & 50.2 & 48.5 & 31.4 \\
\hline
Car-B& {\bf 13.7} & 12.6 & 10.2 & 8.1 & 6.3 & {\bf 87.3} & 55.1 & 71.2 & 48.0 & 37.3 \\
\hline
Car-C& {\bf 12.8} & 12.6 & 10.2 & 8.9 & 5.5 & {\bf 54.2} & 17.0 & 48.1 & 31.0 & 24.6 \\
\hline
Car-D& 14.5 & {\bf 14.6} & 10.0 & 8.7 & 6.8& {\bf 35.6} & 22.3 & 24.7 & 21.2 & 20.2 \\
\hline
Car-E& {\bf 7.4} & 5.4 & 3.5 & 4.1 & 2.7& {\bf 23.5} & 5.7 & 10.9 & 13.0 & 8.0 \\
\hline
Car-F& 28.6 & {\bf 29.7} & 29.1 & 19.5 & 18.3&55.3 & 13.6 & {\bf 59.1} & 28.9 & 35.4 \\ 
\hline
Car-G& {\bf 15.1} & 14.8 & 13.1 & 10.1 & 9.1& {\bf 42.7} & 22.3 & 34.7 & 30.7 & 27.2 \\
\hline
Car-H& {\bf 12.7} & 12.3 & 9.9 & 8.3 & 5.8 & {\bf 93.1} & 50.4 & 47.7 & 74.6 & 33.5 \\
\hline
Car-I& {\bf 18.6} & 14.4 & 12.9 & 6.4 & 8.6& {\bf 13.6} & 4.8 & 9.3 & 4.3 & 8.2 \\
\hline
Car-J& {\bf 18.9} & 17.6 & 13.0 & 11.0 & 8.2& {\bf 114.1} & 72.0 & 113.2 & 78.2 & 49.5 \\
\hline
Car-K& {\bf 19.9} & 18.9 & 13.6 & 13.3 & 7.7& {\bf 26.9} & 9.8 & 15.9 & 11.4 & 10.6 \\
\hline
Car-L& 18.7 & {\bf 20.8} & 14.6 & 13.3 & 8.4&  {\bf 82.3} & 54.4 & 61.1 & 56.2 & 39.5 \\\hline
Car-M& 51.7 & {\bf 58.2} & 57.5 & 42.1 & 50.4&  {31.3} & 7.3 & 23.9 & 13.7 & 30.4 \\\hline
Car-N& 12.3 & {\bf 17.2} & 10.6 & 8.5 & 5.2& {\bf 33.0} & 27.6 & 29.4 & 22.6 & 15.4\\\hline
Car-O& 49.5 & {\bf 68.5} & 62.0 & 52.2 & 47.6& 74.9 & 56.1 & {\bf 85.0} & 72.0 & 50.2\\ \hline
Car-P& {\bf 23.4} & 20.2 & 17.0 & 13.8 & 9.6 &{\bf 74.4} & 38.1 & 66.2 & 26.1 & 43.9 \\\hline
Car-Q& 30.5 & {\bf 32.6} & 30.7 & 21.5 & 17.4 & {\bf 220.4} & 126.6 & 219.1 & 166.8 & 125.5 \\\hline
\hline \hline 
\end{tabular}
\end{center}
\caption{Average $M_{score}(5)$ and $ResM_5$ values for different sequences in 
the PROST, VoT, Cehovin and KITTI sequences.}
\label{tab:scoreNmatches}
\end{table*}

We further demonstrate CoMIC's effectiveness in matching points on real-world vehicle sequences (KITTI)
and popular datasets (PROST, VoT and Cehovin) that have changes in the background.
We present results in terms of $M_{score}(5)$ and $ResM_5$ 
in Table\ref{tab:scoreNmatches}. 
We observe that CoMIC has more points in almost every sequence and returns the best or second best $M_{score}$ 
in most cases closely followed by Hessian and Harris.
These results are especially interesting due to the popular use of point tracking and SfM 
in Vehicle tracking in recent times\cite{forster:2014,song:2013,badino:2013}

\subsection{Discussion}
Point matching is used in a host of 3D applications.
The seminal work by Intille and Bobick\cite{Bobick:1999} uses keypoint matches in a DP-based Stereo matching.
Point tracking is used extensively in Visual 
Odometry\cite{nister:2004,agrawal:2008,forster:2014,song:2013,badino:2013}, SfM from 
video\cite{zhang:2010} and SLAM\cite{klein:2008} where the object or vehicle may move against different backgrounds. 
Even after several years work in Feature Tracking, Kanade-Lucas-Tomasi (KLT) is still the best
algorithm to track points across frames\cite{Baker:2004}. 
When geometric distortion is present, image pyramids\cite{Bouguet:2000} 
or an affine motion models are used\cite{shi:1994}.
While KLT traditionally uses Harris points, FAST features have been used more recently for much higher speed.
Apart from the problems in Harris points, the gradient descent used in KLT will have errors when an object with wired
frames such as a cycle wheel moves in a changing background.
For these reasons many algorithms that use KLT\cite{ali:2013,dorini:2011,lourencco:2012} such as SfM from videos
are unable to use boundary points\cite{zhang:2010}.
CoMIC features integrated into these applications can improve the point detector module 
and thereby improve the overall performance of such systems.
If one could identify the boundary regions (perhaps in an adaptive or probabilistic process), then 
one could also consider using different detectors in the boundary and non-boundary regions or a combination of them
for optimal results.
Hessian, for instance, is a blob detector that has physically different and complementary points that can be used along
with CoMIC to yield best results at both the boundary and non-boundary regions.

\section{Conclusion}
We have presented an iso-curve based method 
that combines contour and appearance information for corner detection and matching. Points on the object 
boundary are detected and matched consistently 
even in changing backgrounds. This is shown in experiments where they perform 
better than the state-of-the-art detectors 
at the object boundaries and comparably at internal regions leading to an overall improvement in performance
for matching the full object.
% for matching the full object.
It yields a sufficient number of stable points on the object that can be used as part of
algorithms for SfM in video sequences, Visual Odometry, stereo etc.
There are several avenues for future work, where CoMIC can be used with more sophisticated motion models
 in a complete tracking setup. Scale and affine invariant extensions can be built based on the 
curve or intensity information
inside, or both.

{\small
\bibliographystyle{ieee}
\bibliography{CoMIC.bib}
}

\end{document}